\documentclass[conference]{IEEEtran}
\IEEEoverridecommandlockouts
\usepackage{cite}
\usepackage{amsmath,amssymb,amsfonts}
\usepackage{algorithmic}
\usepackage{graphicx}
\usepackage{textcomp}
\usepackage{xcolor}
\usepackage{stfloats} 
\usepackage{balance}
\usepackage[colorlinks,citecolor=blue]{hyperref}
\usepackage[margin=19.1mm]{geometry} 

\def\BibTeX{{\rm B\kern-.05em{\sc i\kern-.025em b}\kern-.08em
    T\kern-.1667em\lower.7ex\hbox{E}\kern-.125emX}}
\begin{document}

\title{Adaptive Torque Control of Exoskeletons under Spasticity Conditions via Reinforcement Learning
\thanks{A.C.S. and D.R.C. are funded by the RL-EXO project and 2nd Body project, MOMENTUM program, Spanish National Research Council.}
}

\author{\IEEEauthorblockN{Andrés Chavarrías}
\IEEEauthorblockA{\textit{Neuro AI and Robotics Group}\\
\textit{Cajal International Neuroscience Center}\\
\textit{Spanish National Research Council}}
\IEEEauthorblockA{\textit{Rey Juan Carlos University (URJC)}\\
Madrid, Spain\\
0000-0001-7861-5951}
\and
\IEEEauthorblockN{David Rodriguez-Cianca}
\IEEEauthorblockA{\textit{Neuro AI and Robotics Group}\\
\textit{Cajal International Neuroscience Center}\\
\textit{Spanish National Research Council}\\
Madrid, Spain \\
0000-0002-9503-8508}
\and
\IEEEauthorblockN{Pablo Lanillos}
\IEEEauthorblockA{\textit{Neuro AI and Robotics Group}\\
\textit{Cajal International Neuroscience Center}\\
\textit{Spanish National Research Council}\\
Madrid, Spain \\
0000-0001-9154-0798}
}

\maketitle

\begin{abstract}
    Spasticity is a common movement disorder symptom in individuals with cerebral palsy, hereditary spastic paraplegia, spinal cord injury and stroke, being one of the most disabling features in the progression of these diseases. Despite the potential benefit of using wearable robots to treat spasticity, their use is not currently recommended to subjects with a level of spasticity above ${1^+}$ on the Modified Ashworth Scale. The varying dynamics of this velocity-dependent tonic stretch reflex make it difficult to deploy safe personalized controllers. Here, we describe a novel adaptive torque controller via deep reinforcement learning (RL) for a knee exoskeleton under joint spasticity conditions, which accounts for task performance and interaction forces reduction. To train the RL agent, we developed a digital twin, including a musculoskeletal-exoskeleton system with joint misalignment and a differentiable spastic reflexes model for the muscles activation. Results for a simulated knee extension movement showed that the agent learns to control the exoskeleton for individuals with different levels of spasticity. The proposed controller was able to reduce maximum torques applied to the human joint under spastic conditions by an average of 10.6\% and decreases the root mean square until the settling time by 8.9\% compared to a conventional compliant controller.
\end{abstract}

\begin{IEEEkeywords}
Wearable Robotics, Spasticity, Movement disorders, Reinforcement Learning, Musculoskeletal systems.
\end{IEEEkeywords}

\section{Introduction}
\label{sec:intro}
Spasticity is a neurological impairment consequence of an upper motor neuron syndrome commonly seen in neurodegenerative movement disorders associated with damage to the brain, spinal cord and motor nerves. It often manifests as one of the initial motor symptons in conditions such as cerebral palsy, hereditary spastic paraplegia, spinal cord injury or stroke, and is one of the most disabling features as these diseases progress~\cite{hanssen_contribution_2021, billington_spasticity_2022, lackritz_effect_2021}. This phenomenon was defined by James Lance in 1980 as ``a motor disorder characterized by a velocity-dependent increase in tonic stretch reflexes with exaggerated tendon jerks, resulting from hyperexcitability of the stretch reflex"~\cite{lance1980pathophysiology}. Users with joint spasticity often experience posture and movement abnormalities due to increased muscle tone and persistent primitive reflexes, leading to limitations in walking speed, balance, and specific movements. Within these mobility restrictions, the knee joint presents particular complications due to its crucial role in gait mechanics and weight-bearing~\cite{abid_knee_2019}. 

Wearable robots (WRs), such as robotic exoskeletons, have proven to be valuable tools for therapists in rehabilitation and assistive applications. Under spasticity conditions, the use of WRs has shown that intensive early rehabilitation can potentially prevent the later onset of spasticity, revealing intra-session changes in muscle resistance to passive extension in elbows~\cite{crea_phase-ii_2017} and limb joint~\cite{hui_efficacy_2024}. In addition, it has been shown that the use of exoskeletons in spastic patients during gait has an important influence on the distribution of plantar pressure~\cite{cumplido-trasmonte_modularity_2024}. Despite the potential benefit on using WRs to treat spasticity, their use is not currently recommended to subjects with a level of spasticity above ${1^+}$ in the Modified Ashworth Scale (MAS)~\cite{bohannon_interrater_nodate}, from a maximum level of four, due to safety restrictions. Indeed, the use of WRs under spastic conditions presents several challenges. Conversely to assistance in healthy subjects~\cite{luo_experiment-free_2024}, the primary goal of WRs in movement disorders is not performance (e.g., to speed-up the movement) but to ensure safety (e.g., reducing undesired interaction forces) during the recovery or assistance period. Hence, adaptive control of WRs is an essential feature to ensure safety under different users and scenarios. 

%

\subsection{Related works in WRs control}
While there is wide research in WR control, most current market-available solutions are primarily open-loop driven by PID low-level controllers. In essence, these approaches force the leg to follow a specific trajectory without adapting to user or pathology-specific needs. Thus, these usually need to be calibrated for each subject, implying the need for technicians and experts during rehabilitation, and do not adhere to any complex safety restrictions and physical interactions between the robot, the human musculo-skeletal (MS) system and the environment~\cite{bessler_safety_2021}. 
Model-based controllers are also investigated~\cite{rodriguez-fernandez_systematic_2021}, e.g., model predictive control for prosthetics~\cite{sartori_robust_2018}, but require complex dynamic models that are prone to changes due to user-dependent characteristics and errors due to unknown and changing environmental conditions.

\subsection{Learning to control exoskeletons under movement disorders}
Particularly in the WRs field, where the exoskeleton needs to adapt to both human interactions and the environment, the use of learning-based control algorithms could be greatly beneficial. Among these, RL~\cite{sutton_reinforcement_2018}---learning to interact with the environment by maximizing the expected future reward---appears as one of the most promising approaches given the recent results in assisting healthy users to reduce energy consumption during walking~\cite{luo_experiment-free_2024}. 
Still, the use of RL in WRs, particularly in neurorehabilitation is currently in an early research stay---see~\cite{chavarrias_rl_2024} for a review. Few works focused on enhancing gait rehabilitation but oversimplified the MS-exo system (e.g., joint misalignments) and only addressed muscles weakness pathologies. Besides, RL has been also used for optimizing the parameters of compliant controllers ~\cite{khan_reinforcement_2019}.
RL control under movement disorders, such as spasticity, has not been properly investigated, while being a potential solution due to the challenging nature of the MS changing dynamics, its heterogeneity and safety requirements.

\subsection{Contribution}
This work presents a novel adaptive torque controller via Reinforcement Learning (RL) for a knee exoskeleton (Exo) under joint spasticity conditions. This is, learning to control exoskeletons when the dynamics of the human MS system change depending on its velocity and position, while reducing undesired interaction forces to promote user's safety. This is challenging as the controller should be adjusted for each spasticity level and subject. This work contributes with: $i$) a digital twin of the Human-Exo system that accounts for joint misalignments and relative human-exoskeleton displacements due to soft tissues; $ii$) A differentiable model of different levels of spasticity;  $iii$) an adaptive WR torque controller, providing improved compliant control for different levels of spasticity compared to a standard PID torque controller. 

\section{Methods}
\label{sec:methods}
\begin{figure}[htbp!]
\centerline{\includegraphics[width=8.7cm]{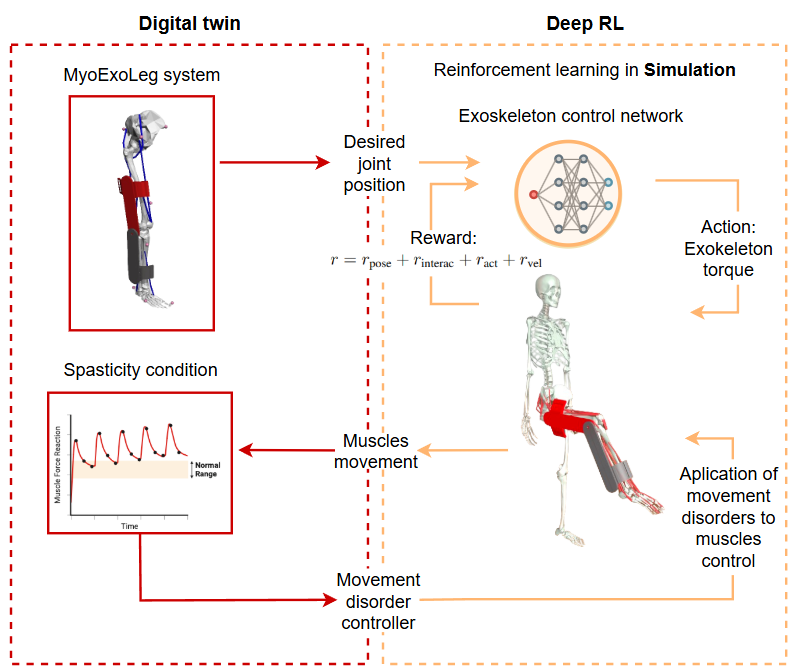}}
\caption{RL Exo Pipeline. Digital twin development with MyoexoLeg system and spastic model in a red box at left. Deep RL architecture in simulation with RL algorithm and environment in orange box at rigth. The interactions between the digital twin and Deep RL architecture are presented at the middle.}
\label{fig:RLexo_pipeline}
\end{figure}

We developed musculoskeletal spasticity models within a digital human-exoskeleton environment and implemented a Deep RL architecture for training and testing an end-to-end torque controller for a knee exoskeleton in the presence of different levels of knee spasticity---Fig. \ref{fig:RLexo_pipeline}. The digital twin includes: the human MS coupled with a knee exoskeleton, incorporating realistic MS-Exo coupling able to account for the presence of joint misalignments (Sec.~\ref{sec:methods:MSExo}), and spastic reflexes model, with a characterization of several pathological spasticity levels (Sec.~\ref{sec:methods:spascticty}). We also describe the Deep RL architecture, explaining the reward function design and the implemented policy improvement algorithm  (Sec.~\ref{sec:methods:dRL}). Finally, we introduce the experimental setup and the simulation framework (Sec.~\ref{sec:methods:experiment}). 

\subsection{Human MS system with Exo coupling}
\label{sec:methods:MSExo}
\begin{figure}[htbp!]
\centerline{\includegraphics[width=8cm]{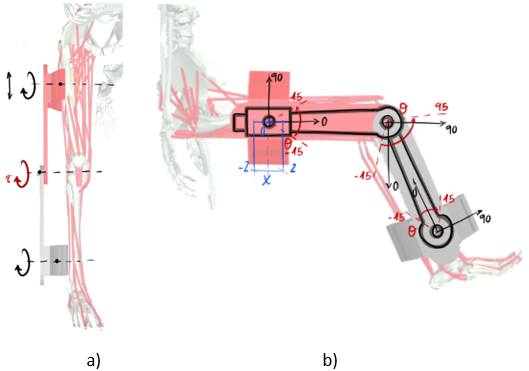}}
\caption{Knee-Exo coupling model with the upper Exo in red and lower Exo in grey. a) Front view of the Human-Exo attachment, including passive joints in black and actuated joints in red. b) Lateral view of the Slider-Crank mechanism designed to simulate a soft knee-Exo coupling, with angular joints ranges for thigh's and shank's cranks in red and thigh's slider range in blue.}
\label{fig:model}
\end{figure}

We modeled an adult of 1,75 m and 70 Kg with a leg exoskeleton (Fig.~\ref{fig:model}), using the Myosuite conversion of Rajagopal's gait model \cite{rajagopal_full-body_2016} as a close reference. The model is a unique stand up body with a fixed 90º right hip flexion, allowing freedom of the knee and ankle joints that includes 40 Hill-based muscles with tendon compliance~\cite{millard_flexing_2013}. Human knee joint has a special feature, it rotation axis varying under the sagittal plane during the flexo-extension movement, which implies  significant challenges in achieving effective coupling with the Exo. To solve these delineations, this MS-Exo union was designed with 4 DOF, 1 DOF for Exo knee joint and 3 DOF simulating a soft coupling with straps.

The MS-Exo coupling (Fig. \ref{fig:model}) was modeled by means of a slider-crank mechanism with two attachment points;  i) the first, on the upper part of the thigh, with two joints, a linear joint with a slide range of \( x_{\text{slider}} \in  [-0.02 m, 0.02 m] \) and a linear friction coefficient of \( \eta = 0.97 \), and a pin joint with a rotation range of \( \theta = \pm15^\circ \), and ii) the second, on the lower limit of the shank with a pin joint and a rotation range of \( \theta = \pm15^\circ \), in order to simulate the relative movement of the exoskeleton due to human soft tissue compliance. The Exo included one active pin joint at the union of the upper Exo (attached on thigh) and the lower Exo (on the shank) to drive the knee joint that included an actuator with a maximum torque of 100 Nm and a joint range of \( \theta_{\text{Torque}} \in  [-15^\circ, 95^\circ] \). 

\subsection{Spastic reflex model}
\label{sec:methods:spascticty}
Joint spasticity is defined as a velocity-dependent muscle tone increase and a muscular activation according to the level of intensity of the muscular reflex and the joint range in the MAS. Here, we provide a spasticity model for muscular activation with a angular velocity parameter~\cite{lee_validation_2004} and a joint range dependency. The parameter of the four spasticity levels  are described in Table: \ref{table:Spasticity levels} following the sign convention presented in Fig. \ref{fig:model}. We merged level 1 with level 1+ due to their close similarities~\cite{pandyan_spasticity_2005} and left out the most severe level (level 4) due to the inability of these subjects to perform any movement without a potentially harmful increase in muscle tone. 

\begin{table}[hbtp!]
\caption{Spasticity Levels and Corresponding Parameters}
\centering
\begin{tabular}{|c|p{4.1cm}|p{3cm}|}
\hline
\textbf{Lv} & \centering\textbf{Muscular activation range limits}& \textbf{Activation coefficient} \\ \hline
0 & Knee joint range ($\theta$): [-45, 95]\(^\circ \) \newline Velocity limit ($v$): [-10, 10] rad/s & Angular Spasticity \(s_{\theta}\): 0 \newline Velocity Spasticity \(s_{v}\): 0 \\ \hline
1 & Knee joint range: ($\theta$): [-35, 80]\(^\circ \) \newline Velocity limit ($v$): [-1, 1] rad/s & Angular Spasticity \(s_{\theta}\): 0.2 \newline Velocity Spasticity \(s_{v}\): 0.2 \\ \hline
2 & Knee joint range ($\theta$): [-30, 75]\(^\circ \) \newline Velocity limit ($v$): [-0.5, 0.5] rad/s & Angular Spasticity \(s_{\theta}\): 0.3 \newline Velocity Spasticity \(s_{v}\): 0.3 \\ \hline
3 & Knee joint range ($\theta$): [-15, 60]\(^\circ \) \newline Velocity limit ($v$): [-0.25, 0.25] rad/s & Angular Spasticity \(s_{\theta}\): 0.3 \newline Velocity Spasticity \(s_{v}\): 0.5  \\ \hline
4 & \multicolumn{2}{|c|}{Not Implemented} \\ \hline
\end{tabular}
\label{table:Spasticity levels}
\end{table}

\subsubsection{Spasticity coefficient to model muscle activation}
 
\begin{figure}[htbp!]
\centerline{\includegraphics[width=8.5cm]{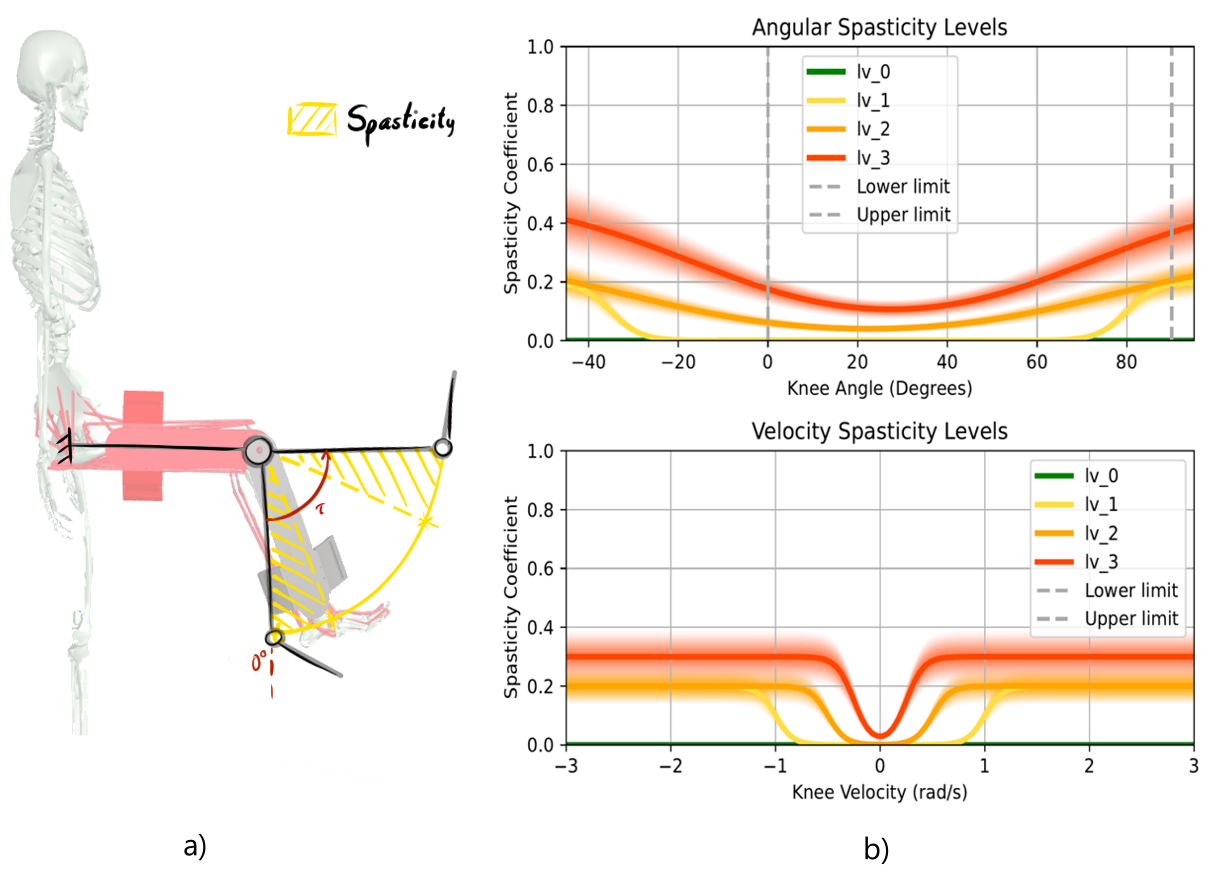}}
\caption{Knee flexion-extension and spasticity model. (a) Spastic regions (yellow), knee joint extension movement (red). (b) Spasticity coefficients as a function of the knee angle (top) and velocity (bottom) for each spasticity level including individual variance to account for inter-subject variability. Level 0 (green), level 1 (yellow), level 2 (orange) and level 3 (red).}
\label{fig:spasticity}
\end{figure}

Muscular activation was implemented as a spasticity coefficient ($SC$), adjusting muscular tone based on angular and velocity readings at the knee joint following Eq. \ref{eq:SCT}. The SC is composed by an angular and velocity coefficient as follows: 
\begin{equation}
SC_{\text{total}} =sc_{\text{angular}}(\theta) + sc_{\text{velocity}}(v)
 \label{eq:SCT}
\end{equation}
Both coefficient functions are defined using sigmoid equations across thresholds (differentiable model) to avoid non-linear transitions and creating spastic reflexes that increase or decrease around specified angle and velocity limits (See table. \ref{table:Spasticity levels} and Fig.~\ref{fig:spasticity}). The appearance and the intensity of spasticity for each individual was modeled by a transition slope $k$ [0, 0.5, 0.2, 0.1] depending on the spasticity level. This slope allows modelling the appearance of spasticity by a muscular activation value in the last part of the joint range for the level 1 and the progressive and maintained behaviour of the upper levels 2 and 3 (Fig.~\ref{fig:spasticity}). The SC equation given the sigmoid adjustable slope $k$ and \(\beta,\beta_0 \) the joint position and velocity range limits results in: 
\begin{equation} 
 \phi(k,\beta,\beta_0) = \frac{1}{1 + e^{k  (\beta - \beta_0)}}
\end{equation}

We define:

\paragraph{Angle-based spasticity coefficient} To account for the angle-dependent nature of spasticity, we added a term that considers both flexion and extension limits for the appearance of spasticity depending on the severity level $sc_{\text{a}}$:
\begin{equation}
sc_{\text{a}}(\theta) = s_{\theta\text{Max}}\left[ 1 - (\phi(k_{\text{ang}}, \theta, \theta_{\text{flex}}) - \phi(k_{\text{ang}}, \theta, \theta_{\text{ext}})) \right]
\label{eq:angular}
\end{equation}
where \(s_{\theta\text{Max}} \) is the maximum SC, \( k_{\text{ang}} \) the slope of the sigmoid transition depending spasticity level muscular activation, \( \theta_{\text{flex}} \) and \( \theta_{\text{ext}} \) the angle threshold for flexion and extension respectively.
\newline
\paragraph{Velocity-based spasticity} Analogously, lower $v_{\text{lower}}$ and upper $v_{\text{upper}}$  velocity limits are used in \( sc_{\text{v}} \) as a function of the velocity \( v \):
\begin{equation}
sc_{\text{v}}(v)  \!=\! s_{\text{vMax}}\left[ 1 - (\phi(k_{\text{vel}}, v, v_{\text{lower}}) \!-\! \phi(k_{\text{vel}}, v, v_{\text{upper}})) \right]
\label{eq:velocity}
\end{equation}

\subsubsection{Using the spasticity coefficient in the MS model}
 $SC_{\text{total}}$ represents a normalized measure of the percentage of maximum isometric force generated by each muscle. Consequently, when spastic reflexes are incorporated into the simulation, each muscle produces a spastic force proportional to its maximum isometric capacity. Spastic reflexes were implemented in 11 muscles involved in knee joint flexion-extension movement grouped into two categories,   based on the literature~\cite{deluzio_principal_1997}:  $i$) Flexion-related muscles, including Gastrocnemius Lateral and Gastrocnemius Medial on the Calf, and Semimembranosus, Semitendinosus, Biceps Sural Long and Biceps Sural Short on the Hamstrings; and $ii$) Extension-related muscles, comprising Gracillis on rectus, and Vastus Medialis, Vastus Lateralis, Vastus Intermedius, and Rectus Femoris on the cuadriceps. 
\newline
\subsubsection{Generating individual subjects}
To account for inter-subject variability and ensure the robustness of the model, the parameters for the different levels of table \ref{table:Spasticity levels}, necessary for eq.\ref{eq:angular} and eq.\ref{eq:velocity}, were generated by introducing Gaussian noise \( \eta \)  with a standard deviation of 10\% of the absolute parameter value. This approach enables the simulation to capture natural variations observed across individuals, providing a more realistic representation of subject-specific dynamics.

\subsection{Deep Reinforcement Learning Architecture}
\label{sec:methods:dRL}
To address the challenges of adaptive exoskeleton control in the presence of spasticity, we designed and trained a deep reinforcement learning architecture able to provide torque control commands to the exoskeleton that fulfill the following objectives: 
 
\begin{itemize}
    \item Learning to reach a specific angular joint position.
    \item Reducing undesired Human-Exo interaction forces.
    \item Minimizing joint velocity. 
    \item Ensuring smooth Exo actuation to guarantee user comfort.
\end{itemize}
These control objectives were translated into a reward function with four main terms: 
\begin{equation}
r = r_{\text{pose}} + r_{\text{interac}} + r_{\text{vel}} + r_{\text{act}}
\end{equation}

Pose reward (\( r_{\text{pose}} \)) is based on the position error (\( e_{\text{pose}} \)), using a exponential equation with \(\sigma^2\) as a growth factor and penalized by a linear term. 

\begin{equation}
r_{\text{pose}} = \exp\left(-\frac{e_{\text{pose}}}{2 \cdot \sigma^2}\right) -\|e_{\text{pose}}\|
\end{equation}

Additionally, a cumulative reward of 0.01 is added to \( r_{\text{pose}} \) when these 2 conditions are satisfied:  $i$) \( r_{\text{pose}} < 0.05  \) rads and $ii$) \( knee_{\text{vel}} < 0.05 \) rads/s. 

\( \tau_{\text{exo\_torque}}\) represent the sum of torques applied (exo and spastic reflexes) on the MS knee joint obtained as an output from Myosuite physics simulator. \( r_{\text{interact}} \) modulates the maximum torque applied to the human knee joint with a linear normalization penalty of \( \nabla_{\text{exo\_torque}} \) and a quadratic penalty based on \(  \tau_{\text{threshold}} \) of 75Nm. 

\begin{equation}
r_{\text{inter}} =
\begin{cases} 
\| \nabla_{\text{exo\_torque}} \|^2, & \text{if } \tau_{\text{exo\_torque}} \leq \tau_{\text{threshold}}, \\ 
\| \nabla_{\text{exo\_torque}} \|^2 + \tau_{\text{exo\_torque}}^2, & \text{if } \tau_{\text{exo\_torque}} > \tau_{\text{threshold}}.
\end{cases}
\end{equation}

\( r_{\text{act}} \) and \( r_{\text{vel}} \) use a linear normalization penalty of the gradients \( [\nabla_{\text{exo\_control}}, \nabla_{\text{knee\_vel}}] \) to reduce undesired motor current spikes and promote a smooth control signal:   
\begin{equation}
\begin{aligned}
r_{\text{act}} &= \left\| \nabla_{\text{exo\_control}} \right\|^2 \quad ; \quad 
r_{\text{vel}} = \left\| \nabla_{\text{knee\_vel}} \right\|
\end{aligned}
\end{equation}

Solving the current continuous control problem using RL requires finding the policy that optimizes the maximum expected reward. From the policy optimizers already tested in WRs~\cite{chavarrias_rl_2024} we selected the Soft Actor Critic Algorithm (SAC)~\cite{haarnoja_soft_2018} due to the changing dynamics of the environment and its exploratory properties. SAC is a policy improvement method with an additional policy entropy parameter for data-expensive and complex environments where the agent has to interact and explore a real-world. The SAC model implemented is summarized in table \ref{table:SAC_architecture}

\begin{table}[h]
\caption{Neural Network Architecture for SAC}
\centering
\setlength{\tabcolsep}{3pt}
\begin{tabular}{l c c c}
\hline
\textbf{Component} & \textbf{Layer Type} & \textbf{Input Dim} & \textbf{Output Dim} \\ \hline
\multicolumn{4}{c}{\textbf{Actor Network (Policy)}} \\ \hline
Input & Observation (state) & 93 & 93 \\
Layer 1 & Fully Connected (Dense) & 93 & 64 \\
Layer 2 & Fully Connected (Dense) & 64 & 64 \\
Output Layer & Fully Connected (Dense) & 64 & 1 \\
Activation & GELU & - & Tanh \\ \hline
\multicolumn{4}{c}{\textbf{Critic Networks (Q1 and Q2)}} \\ \hline
Input & Obs(state) + Action & 93 + 1 & 94 \\
Layer 1 & Fully Connected (Dense) & 94 & 64 \\
Layer 2 & Fully Connected (Dense) & 64 & 64 \\
Output Layer & Fully Connected (Dense) & 64 & 1 \\
Activation & GELU & - & Linear \\ \hline
\textbf{Parameters} & \textbf{Value} & \textbf{Parameters} & \textbf{Value} \\ \hline
 Memory buffer  & $5e^{6}$ & Steps & $5e^{7}$ \\
Adam lr & $10^{-4}$ & Tau & 0.05 \\
Batch size & 4096 & Gamma &  0.99 \\ \hline
\end{tabular}
\label{table:SAC_architecture}
\end{table}

\subsection{Simulation setup and training}
\label{sec:methods:experiment}
The experiment was designed for a point to point trajectory, simulating an almost complete extension of the knee joint  (Fig. \ref{fig:spasticity}.a). The movement begins at a 90º knee flexion position up until a total knee extension rotation of 83º to avoid the risk of knee hyperextension. The policy was derived from the best-performing model after training for 50 million steps, corresponding to approximately 10,000 full knee joint extension cycles. Spasticity levels were uniformly randomized to ensure a balanced number of interactions across all four implemented levels. Additionally, at the start of each iteration, spastic reflex parameters were generated according to the level of spasticity using the methodology described earlier, simulating individual-specific conditions. The results from the SAC controller were compared to those of a baseline PID-based compliant torque controller to assess its improvement over a classical control method. It produced torque control commands based on angular position deviations from the target position using a PID controller tunned to a spasticity level of 1 following the Ziegler-Nichols method with \( K_{\text{p}}= 7\), \( K_{\text{i}}= 40 \) and \( K_{\text{d}}= 0.05 \). This comparison aims to demonstrate the adaptive capability of the SAC controller, which employs a single agent, compared to a classical control method optimized for a single spasticity level.

\section{Results}
\label{sec:results}
The dynamic behavior of control strategies is crucial in managing spasticity during motor tasks, particularly as spasticity severity increases. Hence, we evaluated our developed RL-based controller with respect to the PID-based torque controller over 1000 trials in generated subjects with and without spasticity. We studied the evolution of the following metrics: $i$) leg extension performance, using the angular position error; $ii$) smooth behavior, using the control input signal and $iii$) interaction torques, expressed as the sum of the total torque applied at the knee joint, including the Exo actuation and the spastic reflexes.

\subsection{SAC learned behaviours over spasticity levels}
\begin{figure}[h]
\centerline{\includegraphics[width=0.99\linewidth]{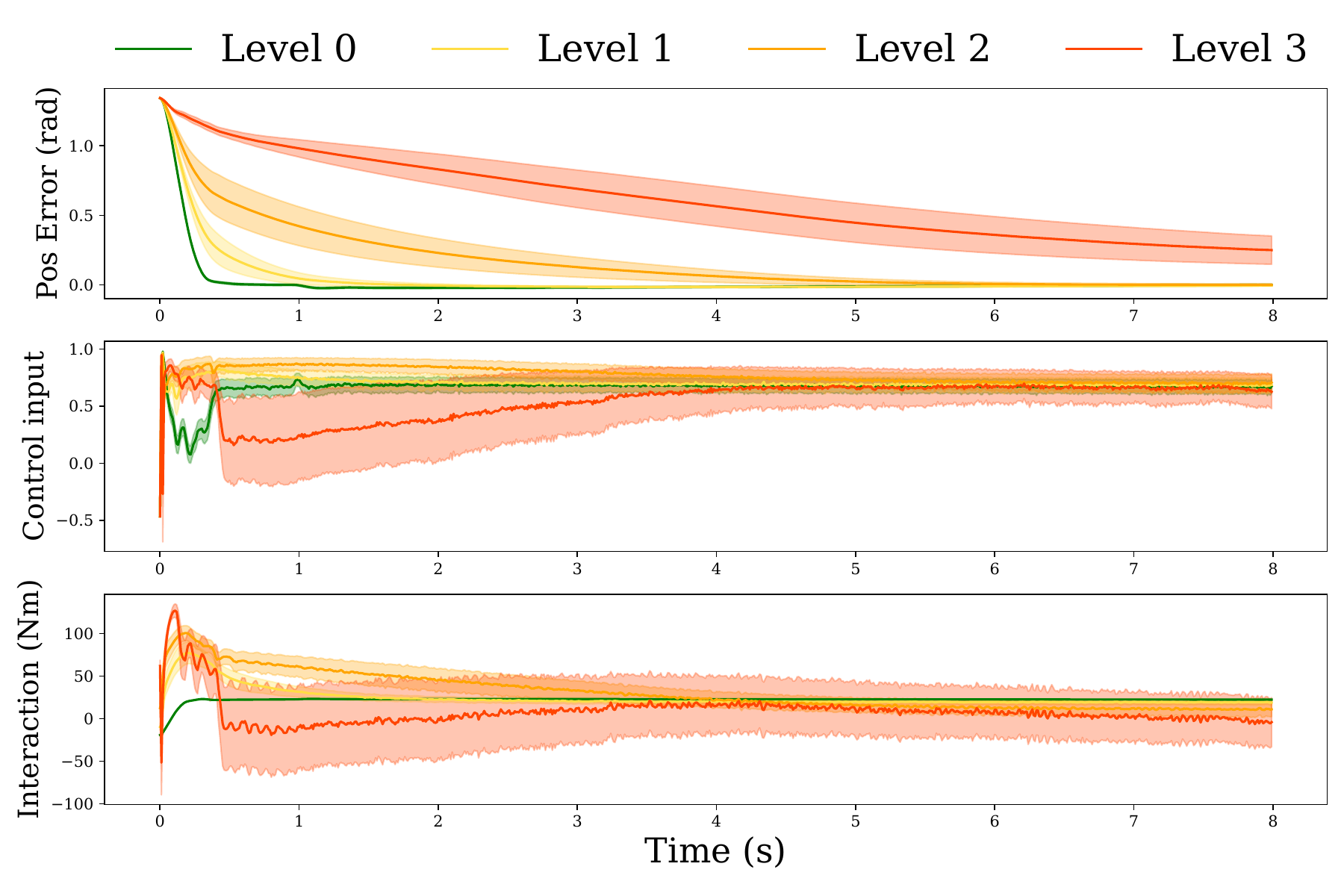}}
\caption{Simulation results of the SAC algorithm for the implemented spasticity levels. In green, yellow, orange and red for level 0 to 3. The upper graph shows the joint position error, the middle graph, the control signal and the lower graphs shows the interaction torques, during 8s of simulation.}
\label{fig:SAC_results}
\end{figure}

Figure~\ref{fig:SAC_results} shows the behavior of the SAC controller for different levels of spasticity, grouped by color. In the upper graph, the angular position error during knee extension, where differences are observed in terms of the evolution and settling time within 2\% of the final value corresponding to 0.26 s, 0.78 s, 3.77 s, 7.94 s for levels 0, 1, 2, and 3, respectively. 

\begin{figure*}[b]
\centerline{\includegraphics[width=0.99\linewidth]{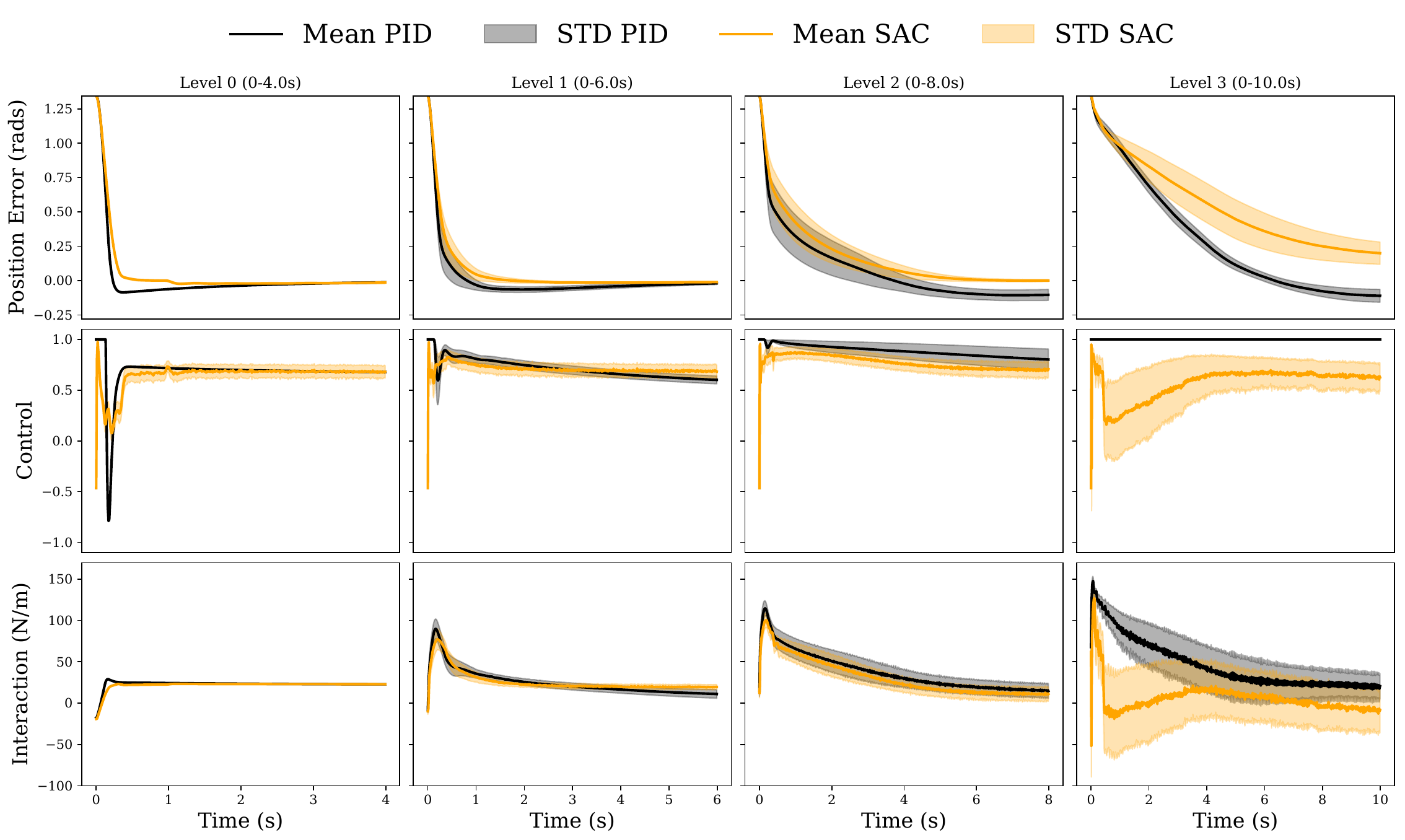}}
\caption{Comparison of PID and SAC algorithms for different levels of spasticity, grouped in columns. In black, PID; in yellow, SAC, representing their mean and standard deviation values. From left to right in columns, the levels of spasticity implemented from a) Level 0 b) Level 1 c) Level 2 d) Level 3. From top to bottom the comparison of a) Position error b) Control signal c) Interaction torque summary exerted by the exoskeleton and spastic reflexes.}
\label{fig:PIDcomparison}
\end{figure*}

As observed, at level 3 the controller is unable to achieve full knee extension, reaching a steady-state angular error of 11.45º. The middle graph corresponds to the exoskeleton control signal, normalized within a control range of [-1, 1] corresponding to [-100, 100] Nm. Its behavior has a closely relation with the observed in the depicting graph, the summary of torques exerted on the knee joint. The results reveals an increase in the exerted torque proportionally to the severity of the spasticity for levels 0, 1 and 2.

\subsection{Comparison with a PID torque controller baseline}

Figure \ref{fig:PIDcomparison} presents a graphical comparison of the RL trained single agent SAC controller over a classical PID torque controller, optimized for a single spasticity level (level 1), for the three metrics discussed earlier (angular position error, control input, and interaction torques), organizing the data by columns corresponding to the four implemented levels of spasticity, with level 0 comparison being the first column. Analysis of these data for the angular position error reveals an overshoot in the PID controller’s performance at levels 0 and 1. At levels 2 and 3, where spastic reflexes are more pronounced, the PID controller achieves full extension with a final error of 6.3º at level 2 and 9.3º at level 3. In contrast, the SAC controller achieves position errors of less than 1º for levels below and equal to level 2 but encounters a superior error of 11.45º at level 3.

The final row of graphs compares the interaction torques between the exoskeleton and the knee joint. A trend similar to the control signal (middle row graphs) is observed, with lower interaction torques values achieved by the SAC controller throughout the simulation period. The difference is more pronounced at spasticity level 3 compared to the other levels. The subsequent results section delves deeper into these interaction torques, analyzing the most representative parameters.

\subsection{Interaction torques analysis}
The aim of the present work was to reduce the MS-Exo interaction forces, thus increasing the safety of the system in spastic conditions. 
\newline
\subsubsection{Reduction of undesired interaction torques during the settling interval}
\begin{figure}[htbp!htbp!]
\centerline{\includegraphics[width=8.5cm]{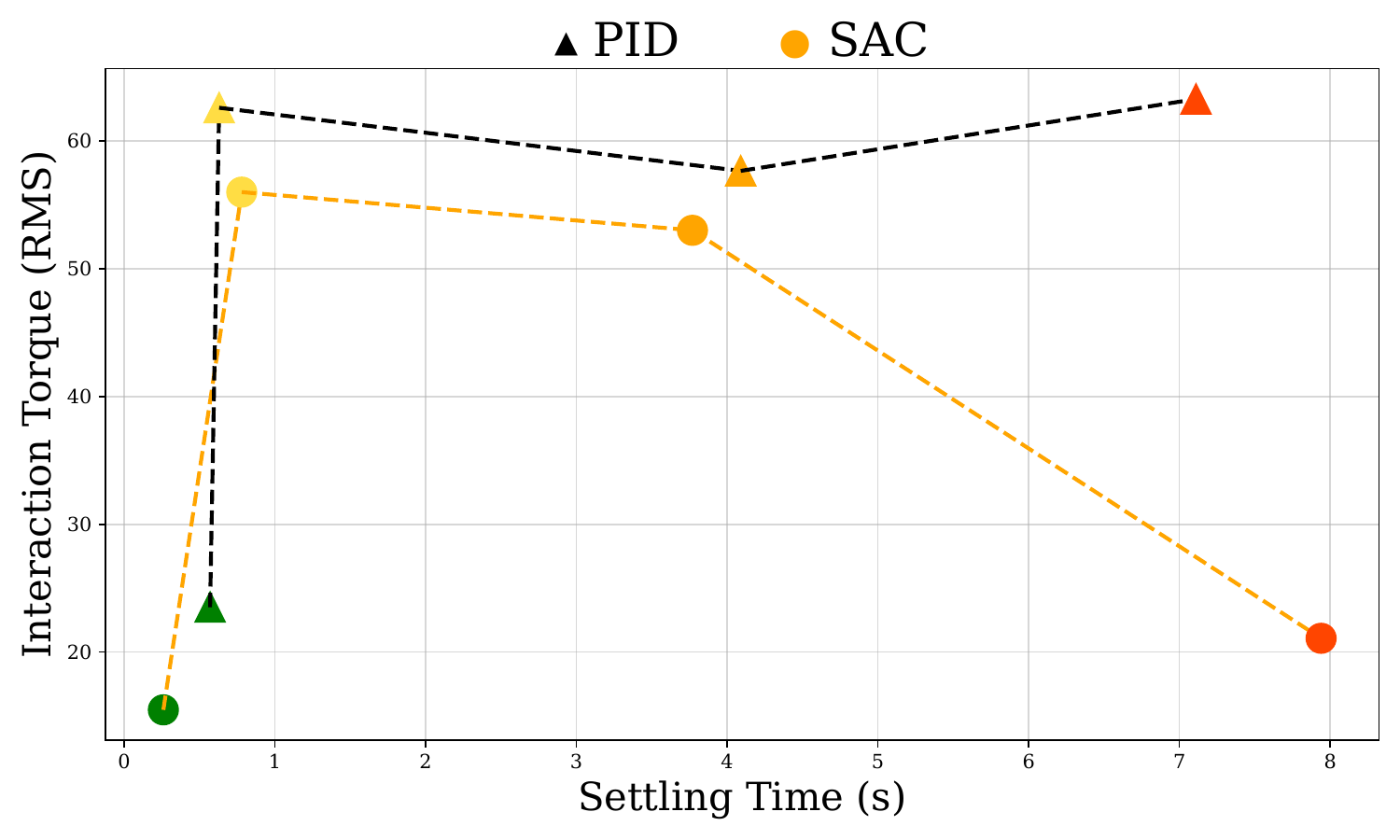}}
\caption{Mean RMS of interaction torques up to the settling time for the implemented spasticity levels. Spasticity levels are color-coded as follows: Level 0 (green), Level 1 (yellow), Level 2 (orange), and Level 3 (red). The control strategies are represented by distinct markers, with circles for the SAC controller and triangles for the PID controller.}
\label{fig:RMS}
\end{figure}

The evolution of the torque exerted by the exoskeleton is analyzed through the Root Mean Square (RMS) value of the interaction torques over the settling time obtained at each level within 2\% of the final value (Fig. \ref{fig:RMS}) for 1000 simulations . The graph demonstrates an increase in settling time corresponding to the severity of spasticity, as well as higher RMS values. However, the SAC controller achieves lower RMS values across all four levels, reducing RMS by 35.3\%, 9.7\%, and 8.1\% in levels 0, 1, and 2, respectively, where full knee extension is achieved.

\subsubsection{Reduction of the maximum torque applied at the knee joint}
\begin{figure}[htbp!htbp!]
\centerline{\includegraphics[width=8.5cm]{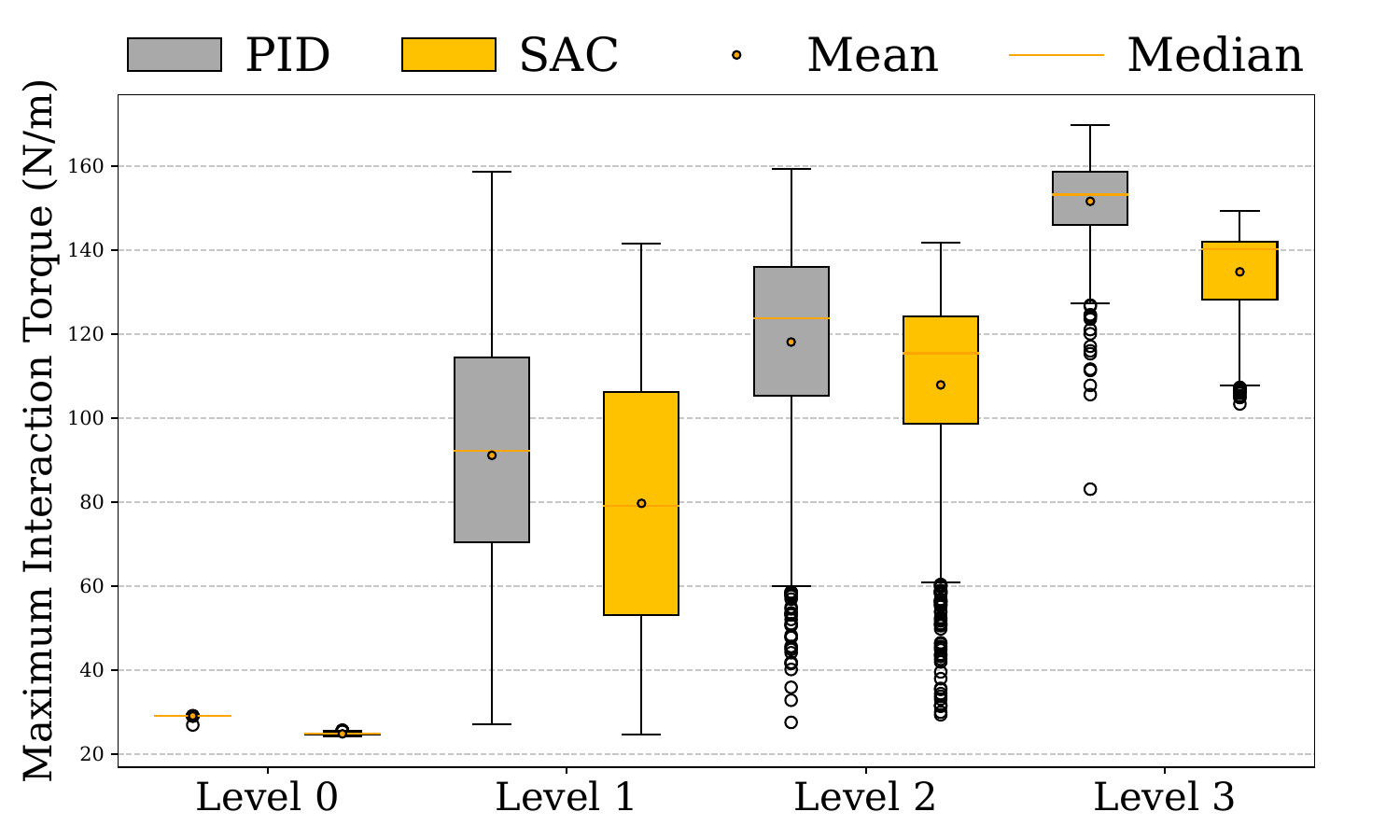}}
\caption{Peak interaction torque for the SAC and PID controllers across the implemented spasticity levels. The results are presented for spasticity levels 0, 1, 2, and 3, with SAC represented in orange and PID in grey. The box plots show the mean (dot), median (line), interquartile range (IQR) (box), and data spread (outliers).}
\label{fig:Box_results}
\end{figure}

Figure \ref{fig:Box_results} compares the maximum torques achieved by the SAC and PID controller for different  spasticity levels. The SAC controller demonstrates a mean peak torque reduction of 10.8±3.4 Nm, corresponding to decreases of 23.6\% at level 0, 12.5\% at level 1, 8.8\% at level 2, and 11.3\% at level 3.

Level 1 exhibits the highest variability in both controllers, with fluctuations in maximum torque reaching a mean of 48 Nm within the Interquartile Range (IQR). Variations of 27.5 Nm and 13 Nm are observed for levels 2 and 3, respectively.

\section{Discussion and Conclusions}
\label{sec:conclusion}
This paper presented an adaptive torque controller for a knee exoskeleton under joint spasticity conditions using RL, offering a suitable approach for wearable robotics for subjects with motor disorders that present muscle position-velocity changing dynamics. The main goal of the approach was to ensure safety (e.g., reducing the amount of undesired interaction forces to complete the task) and provide individualized behavior. 

The results of the proposed method demonstrated the effectiveness of the SAC controller to perform the task and reduce maximum torques applied to the human knee joint under spastic conditions by an average of 10.6\% (Fig. \ref{fig:Box_results}) and decreasing the root mean square (RMS) of the torque until the settling time by 8.9\% (Fig. \ref{fig:RMS}) compared to a conventional PID torque controller. Most importantly, the settling time in these cases shows improved results, with reductions at levels 0 and 2 and similar values at level 1. This indicates that the SAC controller is capable of performing knee extension in less time and with reduced effort. These improvements could translate into enhanced safety during repetitive exercises aimed at spasticity treatment. 

The reduced steady-state interaction torque at the final joint position in high levels of spasticity for the SAC controller avoids the application of high torques over prolonged periods, highlighting the controller priority of user safety (reduced interaction forces) over task performance. This was particularly noticeable at level 3 (Fig. \ref{fig:PIDcomparison}), where the SAC controller prioritized not achieving the target position when the torque required was excessive. Regarding the control input graphs a difference in signal smoothness between the PID and SAC controllers is observed. The SAC controller, driven by the RL algorithm, demonstrates more reactive behavior with a less stable control signal. However, this instability does not propagate to the position response, as the system filters the over-adjustments effectively. Notably, the SAC controller, despite achieving better final position results, maintains a lower control signal than the optimal PID control method across all cases, maintaining a maximum standard deviation below 5\% across randomized spasticity parameters for levels 1 and 2.

Although the evaluation was conducted in a simulated environment, the digital twin model incorporated critical biomechanical characteristics, including joint misalignments and relative displacements of the human exoskeleton due to soft tissue interactions. These considerations enhance the potential for future deployment in physical exoskeleton systems and eventual application in human trials. This physical implementation falls within the scope of the research framework, encompassing the methodologies employed such as domain randomization and physical or non-physical retraining, as well future work aims to extend this approach to address a broader spectrum of motor disorders, such as ataxia and tremor, with the ultimate goal of developing a reinforcement learning-based controller capable of ensuring comprehensive safety and efficacy for individuals with neurodegenerative conditions involving complex and multifaceted movement impairments.

\section*{Environment code availability}
The Myosuite environment used in this study is available at a repository of NAIR Group Github page: \nolinkurl{https://github.com/neuro-ai-robotics/RL_Exo}. 

\section*{Acknowledgment}
We thank Abián Torres, member of NAIR Group, for his insightful help at the algorithmic and data implementation and Pierre Schumacher for technical details with the Myosuite environment.

\bibliographystyle{IEEEtran}
\bibliography{ICORR_Rehabweek}
\end{document}